\documentclass[12pt]{iopart}
\usepackage[utf8]{inputenc}
\usepackage{graphicx}
\usepackage{multirow}
\usepackage[numbers,sort&compress]{natbib}

\expandafter\let\csname equation*\endcsname\relax
\expandafter\let\csname endequation*\endcsname\relax

\usepackage{amsmath}
\usepackage{amssymb}
\usepackage{xcolor}
\usepackage{url}
\usepackage[margin=0.5in]{geometry}
\usepackage{algorithm}
\usepackage[noend]{algpseudocode}
\usepackage{comment}
\usepackage{placeins}
\usepackage{afterpage}
\usepackage{amsthm}
\usepackage{soul}
\usepackage{siunitx}
\usepackage{tabularx}
\usepackage{multirow}
\usepackage{subfigure}
\usepackage{afterpage}
\usepackage{rotating}  
\usepackage{fancyhdr}
\usepackage{tabularx}
\usepackage{pgfplots}
\usepackage{caption}
\usepackage{subcaption}
\usepackage{babel,blindtext}
\usepackage{hyperref}

\usepackage{natbib}
\usepackage{graphicx}

\usepackage{subcaption}

\usepackage{multirow}

\usepackage{float}
\newfloat{algorithm}{t}{lop}

\usepackage{lineno}

\begin{document}

\title[Enhancing Dimensionality Prediction in Hybrid Metal Halides via Feature Engineering and Class-Imbalance Mitigation]{Enhancing Dimensionality Prediction in Hybrid Metal Halides via Feature Engineering and Class-Imbalance Mitigation}
\author{Mariia Karabin$^{1}$, Isaac Armstrong$^2$, Leo Beck$^2$, Paulina Apanel$^2$, Markus Eisenbach$^4$, David B. Mitzi$^3$, Hanna Terletska$^1$, Hendrik Heinz$^2$}
\date{}


\address{$^1$ Department of Physics and Astronomy, Middle Tennessee State University, Murfreesboro, TN, 37132, USA}
\address{$^2$ Department of Chemical and Biological Engineering, University of Colorado, Boulder, CO, 80309, USA}
\address{$^3$ Department of Mechanical Engineering and Materials Science, Duke University, Durham, NC, 27708, USA}
\address{$^4$ National Center for Computational Sciences Division, Oak Ridge National Laboratory, Oak Ridge, TN, 37831, USA}

\begin{abstract}
We present a machine learning framework for predicting the structural dimensionality of hybrid metal halides (HMHs), including organic-inorganic perovskites, using a combination of chemically-informed feature engineering and advanced class-imbalance handling techniques. The dataset, consisting of 494 HMH structures, is highly imbalanced across dimensionality classes (0D, 1D, 2D, 3D), posing significant challenges to predictive modeling. This dataset was later augmented to 1336 via the Synthetic Minority Oversampling Technique (SMOTE) to mitigate the effects of the class imbalance. We developed interaction-based descriptors and integrated them into a multi-stage workflow that combines feature selection, model stacking, and performance optimization to improve dimensionality prediction accuracy. Our approach significantly improves F1-scores for underrepresented classes, achieving robust cross-validation performance across all dimensionalities.

\end{abstract}

%
%
%
%
%

\section{Introduction}

Hybrid metal halides (HMHs) are a promising class of materials for optoelectronic and energy technologies~\cite{priya_2021_accelerated,im_2019_identifying,mitzi_2001_organicinorganic}. Their unique crystal structures, tunable band gaps, and defect tolerance have contributed to the progress in photovoltaics, light-emitting diodes, and photodetectors~\cite{kojima_2009_organometal,stranks_2015_metalhalide,sutherland_2016_perovskite,lin_2018_perovskite,zhang_2021_highperformance}. From a computational standpoint, HMHs are particularly challenging because their hybrid organic-inorganic character combines organic molecules with inorganic frameworks. This computational challenge sometimes requires advanced simulation tools such as the INTERFACE force field (IFF)~\cite{heinz_2013} and motivates data-driven approaches~\cite{wang_2022_datadriven} that can complement first-principles calculations. While 3D perovskites are a well-known subset of HMHs due to their high efficiencies in photovoltaic devices, lower-dimensional structures (0D, 1D, and 2D) may offer improved stability and distinct optoelectronic properties~\cite{kojima_2009_organometal,a2025_best,bayrammuradsaparov_2016_organicinorganic,zhou_2018_low}. Identifying structural dimensionality within the HMH family thus represents a highly important step toward rational materials discovery and design. However, experimental synthesis and characterization of HMHs are resource-intensive, thus motivating the development of computational tools that can accelerate screening and classification.

Machine learning (ML) shows great potential for HMH dimensionality prediction, though faces significant challenges~\cite{xu_2023_small}. First, available HMH datasets are relatively small and far below the volumes typically required for robust ML training~\cite{yao_2024_adapting}. This deficiency increases the risk of overfitting and requires models that can generalize from scarce information. Second, datasets are highly imbalanced, with 2D structures dominating and 0D and 1D being highly underrepresented~\cite{qin_2021_hybridsup3sup}. This imbalance can bias learning algorithms towards majority classes, while maintaining low per-class accuracy for underrepresented classes. Third, the complex interplay of steric, electronic, and structural factors in HMHs requires careful feature design to capture relevant relationships~\cite{wang_2023_feature,wu_2024_universal}. 

To address these challenges, we propose an ML framework combining interaction-based feature engineering, class imbalance mitigation, and advanced ensemble learning~\cite{overman_2024_iife,suguna_2025_mitigating,mienye_2022_a}. First, we design chemically grounded descriptors that include elemental properties, structural characteristics, and interactions relevant to dimensionality. These descriptors are then used to design interaction-based features to capture more subtle, non-linear relationships. We implement the Synthetic Minority Oversampling Technique (SMOTE)~\cite{fernandez_2018_smote,chawla_2002_smote} to augment minority classes and reduce representational bias. Lastly, we evaluate the predictive accuracy and robustness of ML models, which also include stacking ensembles. This cohesive pipeline enhances predictive per-class accuracy while minimizing potential overfitting.

Our study contributes to the broader discussion of small data machine learning strategies in materials science~\cite{xu_2023_small}. More specifically, how to extract reliable insights from small and imbalanced datasets. Many classes of functional materials suffer from similar data scarcity. This limitation motivates the development of ML strategies that involve data-level, algorithm-level, and strategy-level techniques~\cite{zhang_2018_a,mobarak_2023_scope}. Approaches that succeed in these settings are therefore highly transferable and offer generalizable frameworks for tackling diverse small-data challenges in computational materials research. Similar to recent efforts in algorithmic scalability and transfer learning~\cite{gupta_2021_crossproperty}, our approach demonstrates that thoughtful feature design and data augmentation can provide performance gains comparable to larger datasets. 

This work is organized as follows: Section~\ref{related_work} provides a review of the relevant work on machine learning for hybrid metal halide property prediction and small-data strategies. Section~\ref{methods} describes our methodology, including dataset preparation, model development, feature engineering, and class imbalance mitigation. Section~\ref{results} presents the results of our framework, the effects of feature engineering, the impact of SMOTE-based augmentation, and model evaluation. Section~\ref{discussion} discusses the broader implications of our work for HMH discovery and small-data ML strategies, while section~\ref{conclusion} concludes the paper and outlines future research directions.

\section{Related Work}
\label{related_work}

Machine learning for perovskites has emerged as a powerful tool to complement first-principles methods and enable rapid screening of chemical space and property prediction~\cite{tao_2021_machine,kim_2024_machine,chen_2024_application,liu_2023_machine}. Early studies often focused on band gap and stability prediction using composition-based features, geometric or density functional theory (DFT)-derived descriptors~\cite{bartel_2020_a,gpilania_2016_machine,filip_2018_the,lv_2025_thermodynamic}. Other efforts have applied ML to accelerate the discovery of Pb-free perovskite candidates with favorable optoelectronic properties~\cite{im_2019_identifying,lu_2018_accelerated}. These works highlight the ability of ML to handle complex, nonlinear relationships between composition, structure, and functional properties.

Despite these advances, dimensionality prediction in hybrid metal halides (including perovskites) remains relatively unexplored. While several studies have demonstrated the relationship between dimensionality and optoelectronic performance, few ML frameworks explicitly target dimensionality prediction~\cite{yuan_2023_accurate,lyu_2021_predictive}. Yuan et al.~\cite{yuan_2023_accurate} trained ML models on 195 amines to classify low-dimensional organic-inorganic halide perovskites. Lyu et al.~\cite{lyu_2021_predictive} developed ML model to identify 2D perovskites and report steric effect index, eccentricity, largest ring size, and hydrogen-bond donor as key controlling features. One challenge with perovskite dimensionality prediction is that available HMH datasets are typically small and imbalanced. This imbalance can bias model training and limit predictive reliability. Addressing these challenges requires data-level, algorithm-level, and strategy-level techniques tailored for small datasets~\cite{xu_2023_small,yamada_2019_predicting}. Recent progress within small-data strategies also explores the importance of domain-informed descriptors and balanced sampling~\cite{zong_2025_recent}. Techniques such as shotgun transfer learning~\cite{yamada_2019_predicting} and active learning have been demonstrated in broader materials contexts, showing that careful use of limited data can rival results from larger datasets.  

Ensemble-based models~\cite{suguna_2025_mitigating,mienye_2022_a,sagi_2018_ensemble,liorrokach_2009_ensemblebased,rpolikar_2006_ensemble,ali_2015_ensemble} have been shown to be effective in capturing nonlinear feature interactions in small, noisy datasets while maintaining interpretability. Feature importance analysis in these models further provides chemical insight, linking descriptors to structural outcomes~\cite{wang_2023_feature,akbar_2025_analysis,li_2021_improved,anand_2022_topological}. These methods complement deep learning approaches~\cite{maganaie_2022_ensemble} such as graph neural networks, but with the added advantage of interpretability. Such interpretability is important for advancing HMH dimensionality prediction, ensuring that ML not only improves classification accuracy but also contributes to understanding the underlying chemistry~\cite{wu_2024_universal}. Ensemble models such as random forests and gradient boosted trees remain workhorses of small materials datasets, frequently outperforming deep learning models in limited data regimes~\cite{grinsztajn_2022_why,wydmaski_2023_hypertab}. The Synthetic Minority Oversampling Technique (SMOTE) has proven to be effective in balancing materials datasets~\cite{elreedy_2023_a,li_2025_an}. Prior studies show that imbalance skews recall, motivating data-level augmentation approaches like those adopted here. The model tends to overfit the dominant class in the dataset, leading to disproportionately low recall for underrepresented classes and a bias in overall predictive performance.

Our study addresses the underexplored dimensionality prediction in HMHs by designing interaction-based features that encode chemical and structural relationships, mitigating dataset imbalance with SMOTE, and using ensemble models to ensure robustness and interpretability. This bridges the perovskite ML research with the broader small-data ML frameworks, positioning HMHs dimensionality prediction as both a materials challenge and a methodological testbed.

\section{Methods}\label{methods}

In this section, we outline the framework, beginning with dataset construction and preprocessing (section~\ref{dataset}), followed by model development (section~\ref{ML_model}) and feature engineering (section~\ref{features}), and conclude with techniques for mitigating class imbalance (section~\ref{class_imb}). Together these steps define a coherent pipeline to enable robust ML predictions on small, imbalanced datasets of hybrid metal halides.

\subsection{Dataset and feature description} \label{dataset}

\begin{figure}
    \centering
    \includegraphics[width=0.85\linewidth]{figures/HMH_figure.JPG}
    
    \caption{\textbf{(a)} $ABX_3$ Classic perovskite crystal structure illustrating general arrangement; \textbf{(b)} Examples of structural differences between hybrid metal halide dimensionality classes.}
    \label{fig:ABX3}
\end{figure}

The dataset used in this study describes 494 hybrid metal halide (HMH) structures, obtained from the HybriD\textsuperscript{3} Perovskite Database produced by Qin et al.~\cite{qin_2021_hybridsup3sup}. Each dataset entry is correspondingly labeled according to its structural dimensionality. HMHs are a large materials family composed of 'A' organic cations that stabilize 'BX' metal-halide polyhedra which consist of group 14 metals (B) and halide anions (X)~\cite{brenner_2016_hybrid}. Conventional ABX\textsubscript{3} hybrid perovskites, of recent solar research interest, are a subset of HMHs (Figure~\ref{fig:ABX3}a). Dependent on the underlying chemistry, these metal-halide octahedra may share corners, edges, or faces, and generally arrange into one or more different dimensionalities: 0D - lacking an extended inorganic framework, 1D - chains, 2D - sheets, and 3D - bulk (Figure~\ref{fig:ABX3}b). Structural dimensionality of HMHs has to-date been difficult to predict a priori, with similar compounds exhibiting different dimensionalities with only minor changes in composition or synthesis environment~\cite{mitzi_2001_organicinorganic,jana_2020_organictoinorganic,lyu_2021_predictive}.

\begin{figure}
    \centering
    \includegraphics[width=0.4\linewidth]{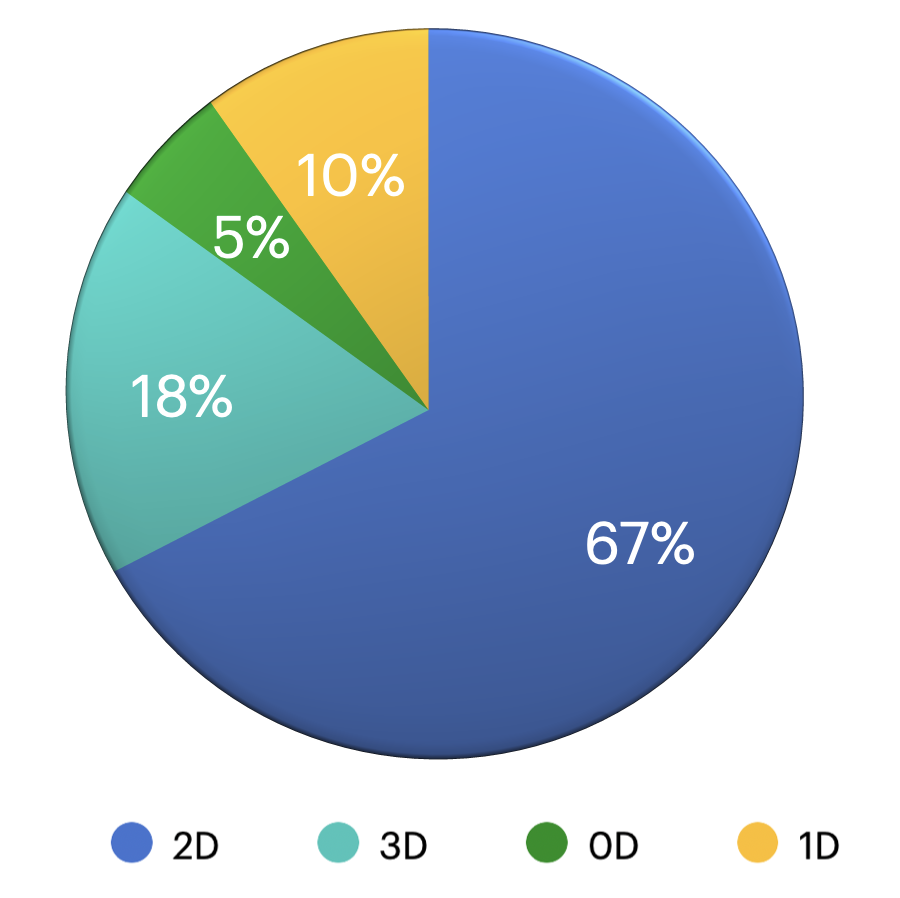}
    \caption{Data distribution of structural dimensionality of hybrid metal halides (HMHs)}
    \label{fig:data_overview}
\end{figure}

 Figure~\ref{fig:data_overview} reflects high data imbalance and prevalence of layered 2D HMHs, which comprise 67\% of the dataset. In contrast, 0D and 1D structures correspond to only 5\% and 10\%, respectively.
In addition to the assigned dimensionality, each entry in the HybriD\textsuperscript{3} dataset contains the complete chemical composition of the HMH. From these compositions, we generated atomic-level, molecular-level, as well as geometric features. These features, together with the interaction-based features described in section~\ref{features}, serve as the input for the machine learning model described in section~\ref{ML_model}.

To start, we derived our 11 original features (Table~\ref{tab:features}) from our dataset in an effort to incorporate chemical effects that could influence dimensionality. This effort was guided by an analysis of our dataset that identified 17 instances of two or more HMHs that changed dimensionality upon inclusion of a different organic cation for the same inorganic lattice composition (i.e. B = Pb, X = I). Due to the higher prevalence of examples of dimensionality change upon replacement of cations, as opposed to swapping out inorganic compositions, we focused our features on traits of organic cations. Geometric dimensions of the cation were an immediate starting point, leading us to define the presence of alkyl chains, aromatic rings, and their relative sizes, generating 6 descriptors. A vast majority of our listed cations consisted of C,H, and N atoms, with N being the most electronegative. As such, we described relative N placements within the cations by simply counting how many 'terminal' N atoms were present (N atoms bonded to only a single non-hydrogen atom). Some entries contained multiple 'A' cations per unit cell of the HMH, the number of which we included. Additionally, some of the HybriD\textsuperscript{3} data had been reported with water present, potentially displacing cations, so the presence of water was also tracked. These features, coupled with the chemical formulas for the organic and inorganic components, gave 11 unique descriptors from which to initialize ML training. Further details on these descriptors, and why they were chosen, can be found in the SI.

\subsection{ML model development and validation}\label{ML_model}

To predict the dimensionality of hybrid metal halides, we employed both single model learners and an ensemble strategy to leverage complementary strengths. The implemented models were selected for their robustness in handling heterogeneous features, non-linear relationships, and imbalanced datasets~\cite{mienye_2022_a,shmuel_2025_a}.

Gradient Boosted Trees (GBT) were used as a core predictive model due to their ability to capture complex non-linear decision boundaries and interactions between features~\cite{bentjac_2020_a}. GBT iteratively fits weak learners (decision trees) to the residual errors of previous trees to reduce bias and variance. This iterative boosting framework is particularly well-suited for tabular scientific datasets, where descriptors exhibit varying degrees of freedom. Hyperparameters such as learning rate,  maximum tree depth, and number of estimators were tuned to balance predictive accuracy with generalization.

In addition to GBT, we have also used several classifiers as base learners in the ensemble. Logistic regression is a linear model that provides interpretable baseline decision boundaries, while Random Forest is a bagging-based ensemble of decision trees that mitigates overfitting. Support Vector Machine (SVM) is effective in high-dimensional feature spaces, particularly for separating overlapping classes. These models capture distinct structural patterns and provide complementary perspectives on the dataset~\cite{li_2021_predicting}.

We implemented a stacking ensemble to integrate these predictive strengths. The base learners generate first-level predictions, which are then used as input features for the meta-learner, GBT. This two-layer architecture makes use of diverse error structures and learns optimized decision rules that outperform an individual model. Stacking has been shown to improve robustness against overfitting and provide better generalization on small, imbalanced datasets~\cite{xu_2023_small,zhang_2018_a}. Our model was trained and validated using stratified k-fold cross-validation to preserve class balance across folds. Performance was evaluated using precision, recall, F1-score, and receiver operating characteristic (ROC) curves and corresponding area under the curve (AUC) scores (section~\ref{results}). The ROC curve shows the model's ability to distinguish between different classes by plotting the true positive rate against the false positive rate. A higher AUC value indicates a stronger capability of the classifier to distinguish between different structural dimensionality classes, with an AUC of 1.0 representing perfect separation.

\subsection{Feature engineering}\label{features}

Our workflow involves interaction-based feature engineering~\cite{overman_2024_iife,verdonck_2021_special}, as the original descriptors alone were insufficient to fully capture the structure-dimensionality relationships of HMH. Given the relatively small size of our dataset (494 compounds), careful feature design was required to maximize the prediction accuracy while minimizing the risk of overfitting. This approach combines chemically motivated descriptors with interaction-based features that encode more subtle relationships, and therefore improve separability among minority classes.

The original 11 descriptors were derived from our dataset (section~\ref{dataset}) and reflect steric hindrance, hydrophilicity, and molecular complexity, all of which highly affect HMH dimensionality. To capture nonlinear effects and coupled contributions, we introduced 11 interaction-based features (Table~\ref{tab:features}). Unlike statistical feature construction, this approach reduces the risk of spurious correlations, which is particularly important in small datasets where ML models are more prone to overfitting~\cite{xu_2023_small}. Moreover, interpretable features allow ML predictions to be validated against existing chemical intuition. 

The pairwise interaction features were generated using pairwise combinations of the original features (e.g., differences, products, and ratios). This encodes competitive or synergistic effects between features that influence the dimensionality of HMHs. Other types of interaction features were generated using domain-informed combinations and non-linear transformations (Table~\ref{tab:features}). 

\begin{table}[h!]
    \centering
    \begin{tabular}{p{3.2cm} p{6.5cm} p{6.0cm}}
    \hline
    \textbf{Feature Type} & \textbf{Feature} & \textbf{Description} \\
    \hline
    Original features &  Organic and inorganic formula units, number of cation rings, cation ring C count, cation ring non-C count, longest alkyl length chain, number of alkyl chain, water presence (boolean), number of terminal nitrogens, longest chain C count, number of same type cations & Captures compositional effects and structural factors that govern steric packaging, hydrogen bonding, and polarity effects influencing dimensionality.\\
    \hline
    Interaction-based features & Total alkyl chain weight, ring length interaction, normalized count of terminal nitrogens per chain length, molecular weight per chain length, cation complexity, nitrogen/ weight ratio, compactness, hydrophilicity index, size complexity & Encode nonlinear or coupled effects between the original features reflecting how chemical interactions and packing affect HMH dimensionality.\\
    \hline
    \end{tabular}
    \caption{Summary of original features and engineered interaction-based features used for HMH dimensionality prediction, along with their description.}
    \label{tab:features}
\end{table}

As shown in sections~\ref{results_features} and ~\ref{class_imbalance}, the engineered interaction features improved F1-scores across all dimensionality classes, particularly in underrepresented categories (0D and 1D).

\subsection{Class imbalance}\label{class_imb}

The original HMH dataset (section~\ref{dataset}) exhibits high class imbalance, with 2D structures being prevalent and 0D and 1D structures highly underrepresented (Figure~\ref{fig:data_overview}). This imbalance presents a significant challenge for developing a machine learning model with high generalizability and prediction accuracy~\cite{jiang_2025_a,basharhamadaubaidan_2025_a,chen_2024_a,zhao_2025_a}. The models trained on the original distribution tend to overfit the majority class (2D) and perform poorly on the minority classes (0D and 1D), as reflected in the low F1-scores (section~\ref{results_features}).

To address this issue, we applied a Synthetic Minority Oversampling Technique (SMOTE)~\cite{chawla_2002_smote,fernandez_2018_smote} to generate additional samples for minority classes. Producing a new synthetic sample requires interpolation between a randomly chosen minority class sample and one of its k-nearest neighbors in feature space:
\[x_{\text{new}} = x_i + \lambda\,(x_{nn} - x_i), \quad\lambda\in[0,1]\]

SMOTE constructs a synthetic data point by taking a weighted combination of two real samples from the same minority class in the descriptor space. This means that each generated sample does not correspond to a physically realized compound, but to an interpolated representation with feature values lying between two chemically plausible points. The algorithm selects k nearest neighbors for each class, ensuring that the synthetic points populate regions of the feature space that were originally sparse. This feature-space interpolation effectively creates pseudo-structures that extend the distribution of the existing samples. The augmented samples maintain chemical realism since they were derived from physically valid feature relationships.

This approach improves model generalization and mitigates overfitting by increasing the diversity of the minority class samples without simply duplicating the existing ones. In our workflow, the augmentation balanced the dataset from its original 494 HMHs samples to a total of 1336 samples, with each dimensionality class containing an equal number of 334 samples per class. This reduces the bias towards the 2D majority class and improves decision boundaries between classes in the multi-class setting, as reflected in the ROC curves and per-class AUC values in section~\ref{class_imbalance}.

\section{Results} \label{results}

In this section, we present the results of our dimensionality prediction framework for HMHs. Section~\ref{results_features} highlights the effect of feature engineering, particularly the introduction of interaction descriptors on classification performance. Section~\ref{class_imbalance} evaluates the impact of class imbalance mitigation using SMOTE and its role in improving prediction for minority classes. Section~\ref{evaluation} includes a detailed analysis of per-class F1-scores as well as cross-validation analysis.

\subsection{Feature engineering} \label{results_features}

To improve the prediction accuracy of the structural dimensionality of HMHs, we developed chemically-informed interaction-based descriptors alongside original compositional features. The initial feature set includes basic atomic and molecular descriptors as well as features derived from them. While these features provided some separation between dimensionality classes, they were not sufficient to capture the subtleties of structural dimensionality patterns corresponding to lower-dimensional HMHs.

As shown in Table~\ref{table:f1_scores_original}, the model trained on the original set of descriptors provides high accuracy for the majority class (2D, F1-score = 0.89), but underperforms significantly on the minority classes. Specifically, F1-score for 0D and 1D structures are 0.29 and 0.36, respectively, showing poor generalization to underrepresented dimensionalities. The support values in Table~\ref{table:f1_scores_original} show the number of test samples for each dimensionality class. These values highlight the dataset imbalance, with 2D structures being the majority class, and provide context for interpreting the class-specific F1-scores. 

To address these limitations, we designed an additional set of descriptors based on key interaction components, such as metal-halide coordination geometries, organic cation steric effects, and measures of structural compactness. These descriptors were designed to better reflect the spatial and chemical connectivity relevant to structural dimensionality.

\begin{table}
\centering
\setlength{\tabcolsep}{12pt}
\begin{tabular}{rrc}
Dimensionality & F-1 score & Support \\
\hline
0 & 0.29 & 5 \\
1 & 0.36 & 5 \\
2 & 0.89 & 67 \\
3 & 0.71 & 22 \\
\hline
\end{tabular}
\caption{Per-class F1-score and support for ML model trained on original descriptors.}
\label{table:f1_scores_original}
\end{table}

\begin{figure}[htbp]
    \centering
    \includegraphics[width=0.8\linewidth]{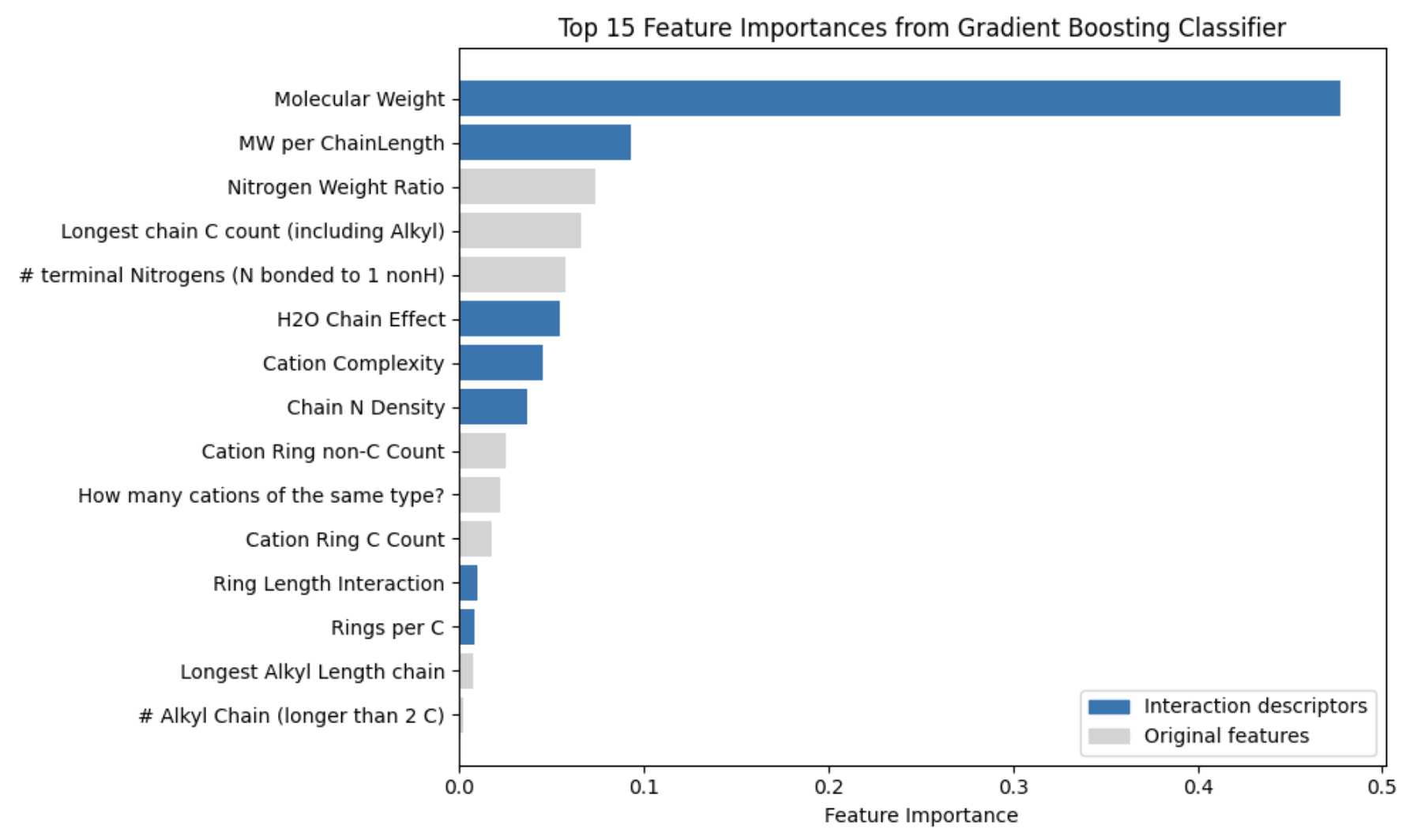}
    \caption{Feature importance analysis of all of the descriptors for HMHs dimensionality prediction.}
    \label{fig:feature_importance}
\end{figure}

To gain insight into the contribution of the individual descriptors we performed feature importance analysis using a gradient boosting classifier trained on the original descriptors as well as interaction-based descriptors. As shown in Figure~\ref{fig:feature_importance}, the most influential descriptors include 7 original features and 8 interaction-based features, most of which relate to the size and chemical composition of the cations. Structural descriptors such as the longest carbon chain, number of terminal nitrogen atoms, and hydrogen bonding also contribute significantly to prediction accuracy. These results highlight the relevance of both steric and electronic features in capturing the nuances relevant to dimensionality prediction.

The interaction descriptors enriched the feature space with the information that is highly important to distinguishing structural dimensionality of HMHs. This refinement was a necessary first step before applying additional class imbalance mitigation techniques described in the next section.

\subsection{Effect of class imbalance mitigation} \label{class_imbalance}

\begin{figure}
    \centering
    \includegraphics[width=0.65\linewidth]{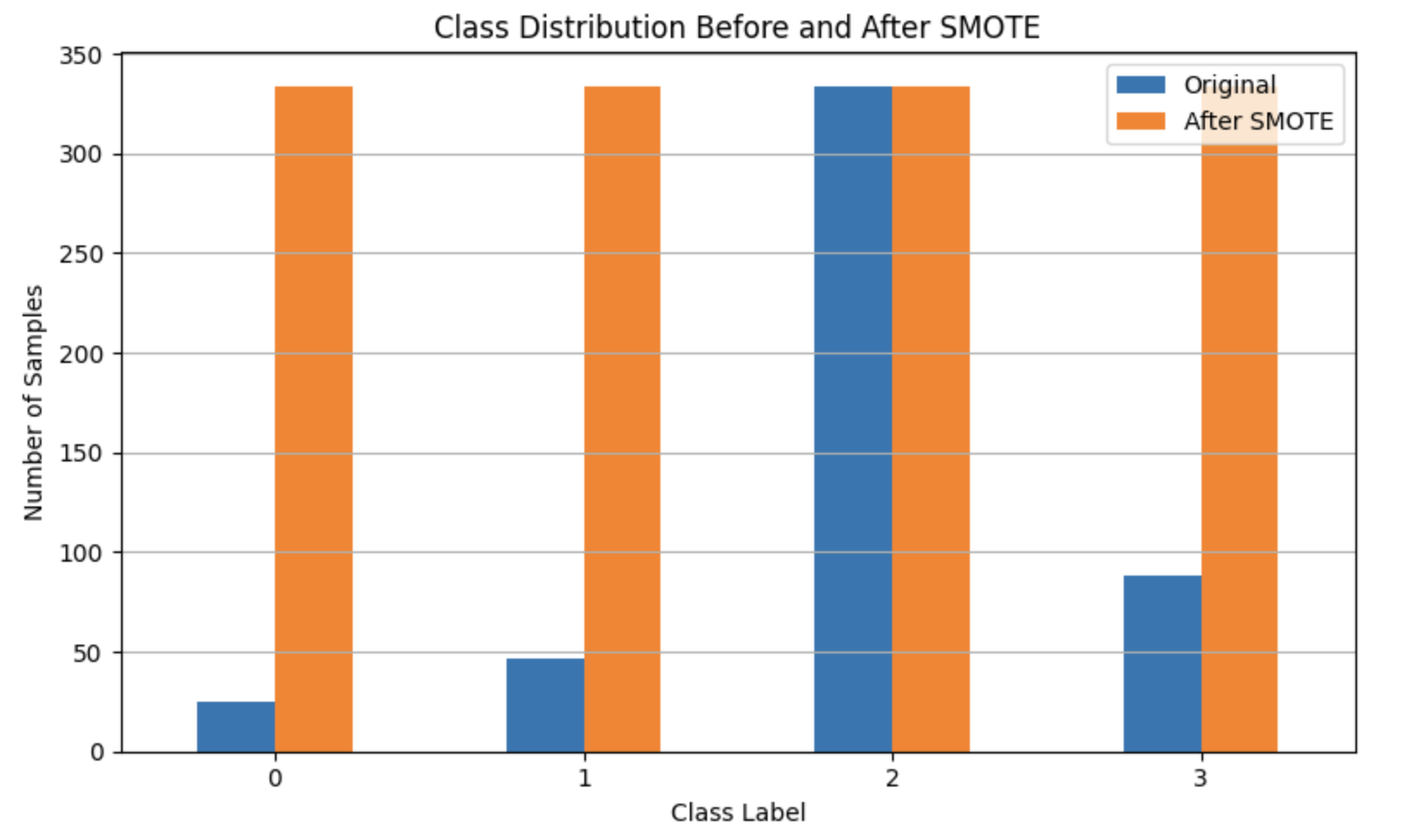}
    \caption{Per-class data distribution before and after SMOTE-based oversampling. The original dataset (blue) shows a strong imbalance favoring 2D structures. After the data augmentation (orange), all four dimensionality classes contain an equal number of samples, enabling better model generalization.}
    \label{fig:smote}
\end{figure}

 The original dataset shows a significant class imbalance, with 2D structures constituting the majority of samples, while 0D and 1D classes are highly underrepresented (Figure~\ref{fig:data_overview}). As demonstrated in Figure~\ref{fig:smote}, this imbalance was addressed using the Synthetic Minority Oversampling Technique (SMOTE), which synthetically generates new samples of the minority classes by interpolating between the existing data points. After applying SMOTE, the dataset was expanded from 494 samples to 1336 samples, with all four dimensionality classes being represented equally.

This resampling strategy led to substantial improvements in classification performance for minority classes. As shown in Table~\ref{table:f1_scores_sampling}, the F1-score for 0D structures increased from 0.29 (see Table~\ref{table:f1_scores_original}) to 0.74, and for 1D structures from 0.36 (Table~\ref{table:f1_scores_original}) to 0.83. The F1-scores for the 2D and 3D majority classes remained high at 0.75 and 0.79, respectively. Support values in Table~\ref{table:f1_scores_sampling} show the number of test samples for each dimensionality class after applying SMOTE oversampling. The balanced support across 0D - 3D classes reflects the effectiveness of data-level augmentation in mitigating class imbalance.

\begin{table}
\centering
\setlength{\tabcolsep}{12pt}
\begin{tabular}{rrc}
Dimensionality & F-1 score & Support \\
\hline
0 & 0.74 & 69 \\
1 & 0.83 & 61 \\
2 & 0.75 & 73 \\
3 & 0.79 & 65 \\
\hline
\end{tabular}
\caption{F1-scores and support after applying interaction-based descriptors and SMOTE oversampling. Minority class performance improves substantially, while maintaining high performance for majority classes.}
\label{table:f1_scores_sampling}
\end{table}

The improved balance in class representation enabled the model to learn more generalizable patterns across all dimensionalities, rather than overfitting to the dominant 2D class. These results highlight the importance of data-level augmentation techniques in small and imbalanced materials datasets.

\subsection{Model evaluation} \label{evaluation}

\begin{figure}
    \centering
    \includegraphics[width=0.55\linewidth]{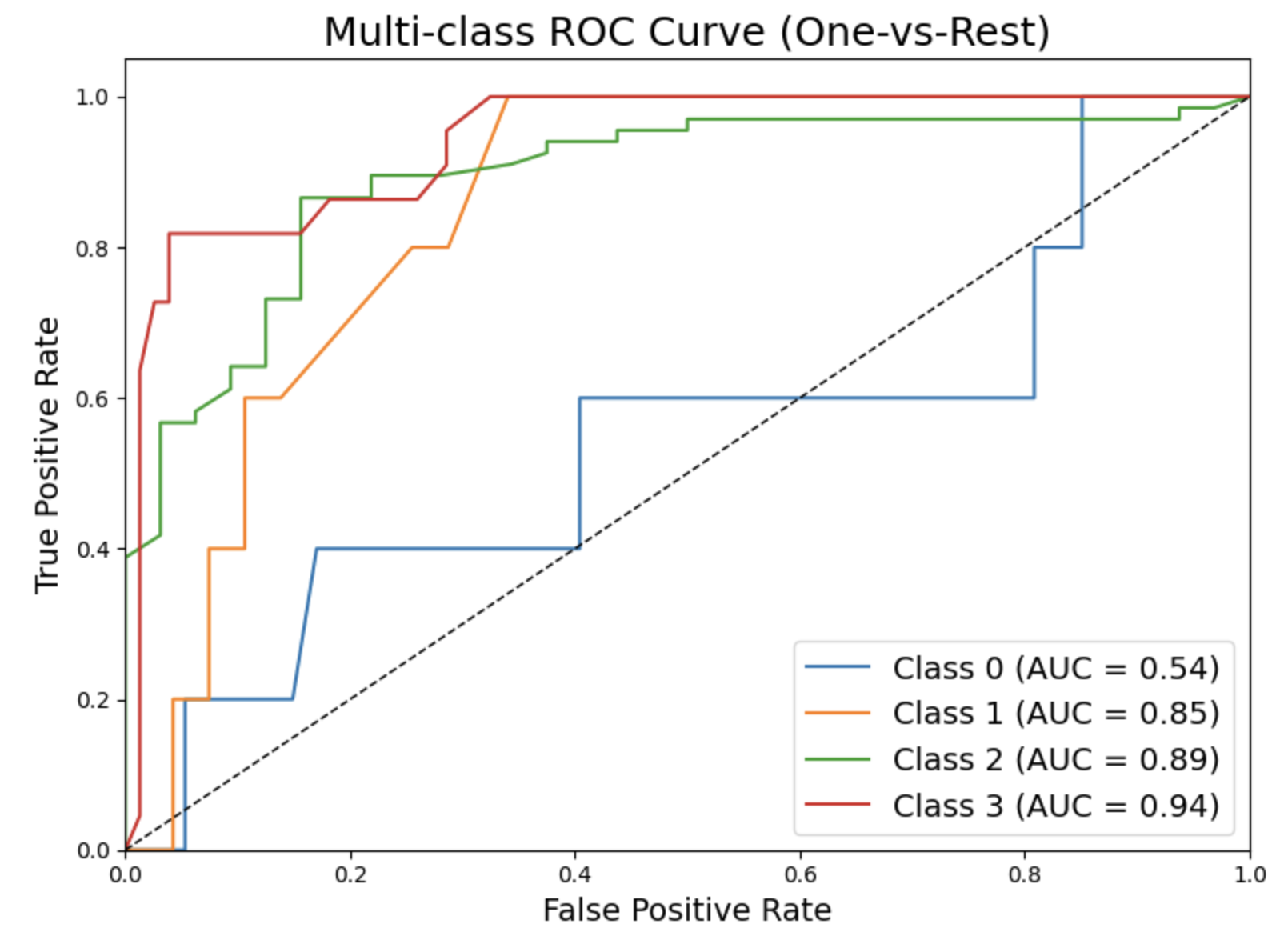}
    \caption{Multi-class ROC curves (one-vs-rest) before SMOTE augmentation. Minority class 0 (0D HMHs) performance is low (AUC=0.54), indicating poor classification due to class imbalance.}
    \label{fig:roc}
\end{figure}

To further evaluate the model performance and prediction accuracy across individual classes, we calculated one-vs-rest receiver operating characteristic (ROC) curves and corresponding area under the curve (AUC) scores. Figure~\ref{fig:roc} shows the ROC curves generated using the original (imbalanced) dataset. The classifier performs well on the majority classes (AUC = 0.89 for class 2 (2D HMHs) and 0.94 for class 3 (3D HMHs)), but fails to accurately distinguish minority classes, with AUC for class 0 (0D HMHs) of only 0.54.

\begin{figure}
    \centering
    \includegraphics[width=0.55\linewidth]{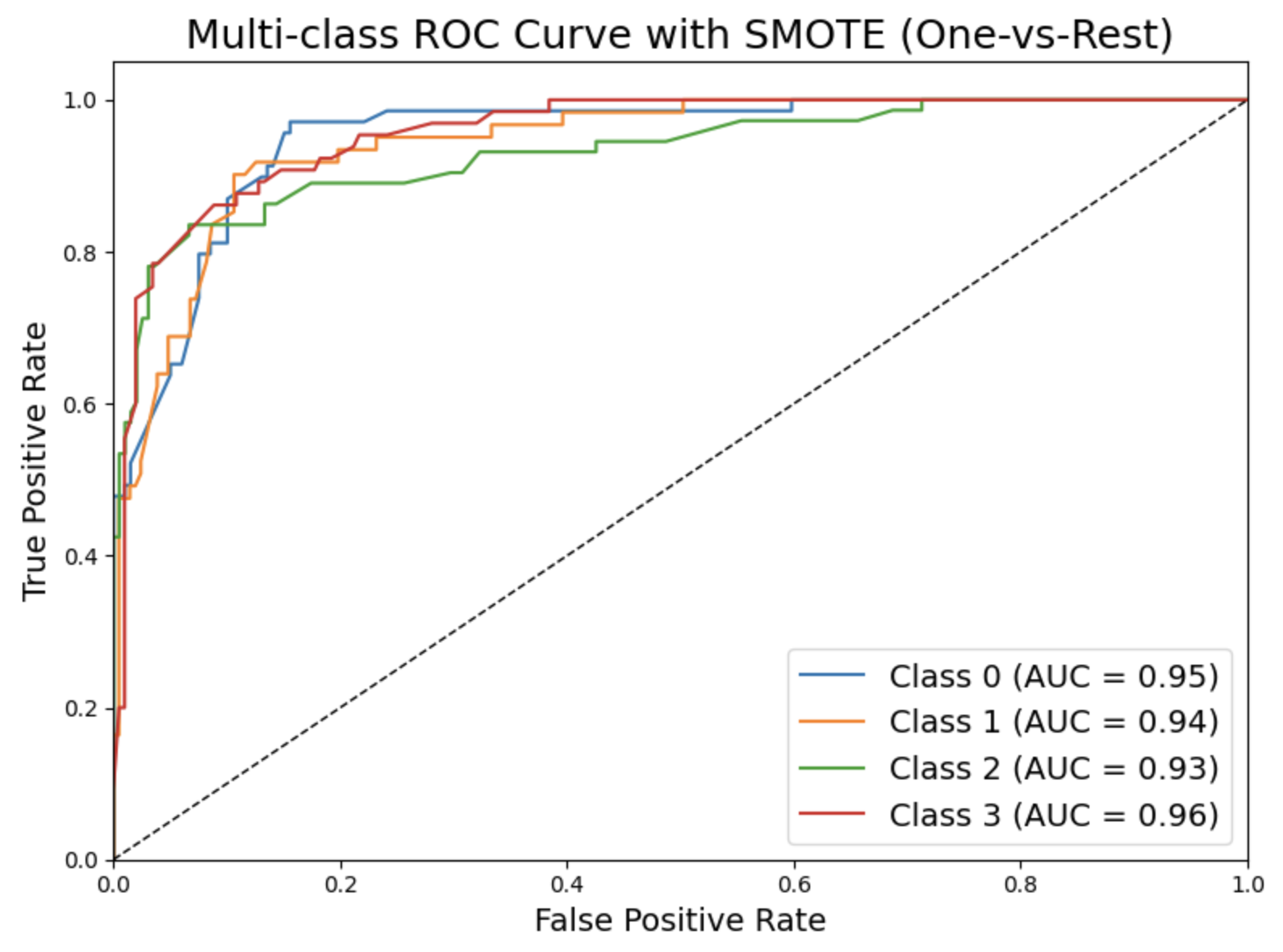}
    \caption{Multi-class ROC curves after SMOTE augmentation. AUC values improve for all classes, particularly for class 0 (AUC=0.95), demonstrating that oversampling significantly enhances classification performance for minority classes.}
    \label{fig:roc_after}
\end{figure}

After applying SMOTE, performance significantly improves across all classes, as shown in Figure~\ref{fig:roc_after}. The AUC for class 0 (0D HMHs) increases from 0.54 to 0.95, and classes 1,2, and 3 also show improvements to 0.94, 0.93, and 0.96, respectively. These results indicate that SMOTE-based augmentation significantly improved class representation as well as the model's ability to accurately distinguish between the dimensionality classes.

\begin{figure}
    \centering
    \includegraphics[width=0.75\linewidth]{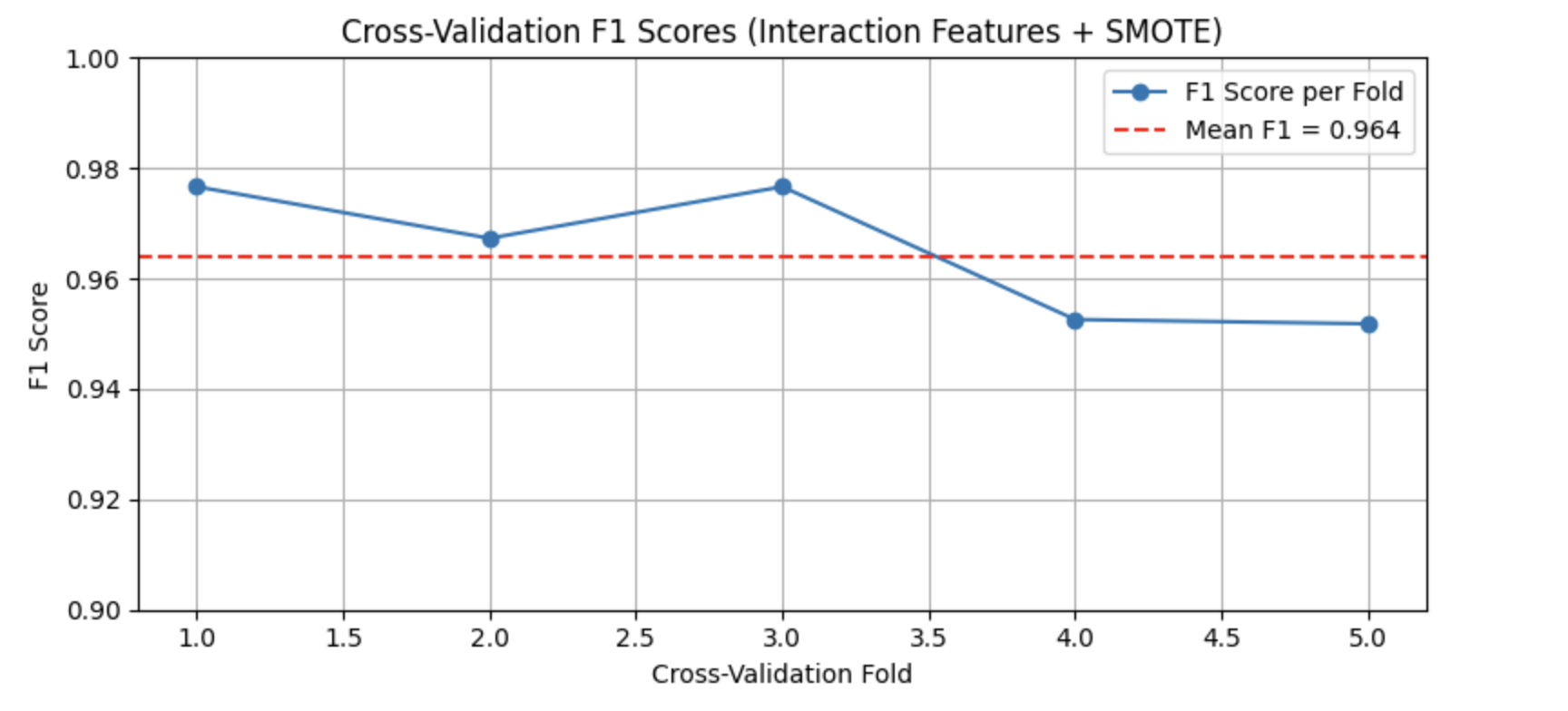}
    \caption{Cross-validation F1-scores across five folds for the model trained on interaction-based descriptors with SMOTE. The model achieves a mean F1-score of 0.964 (red dashed line), with low variance across folds, demonstrating the model's stability (blue line).}
    \label{fig:CV}
\end{figure}

We conducted a 5-fold cross-validation using the final model trained on the interaction-based descriptors and the SMOTE-augmented dataset to assess the robustness and generalization capabilities of our model. As shown in Figure~\ref{fig:CV}, the model exhibits consistently high performance across all folds, with F1-scores ranging from 0.947 to 0.98 and a mean F1-score of 0.964 (red dashed line). The resulting variation between folds corresponds to an uncertainty margin of $\pm$1.3\%, confirming the robustness and reproducibility of the model's predictive performance. This stability across folds suggests that the model is not overfitting to the specific subset of the data and is generalizable across a range of HMH compositions and structural dimensionalities. To evaluate model performance consistently, the dataset was split into training and test subsets using a stratified random split, ensuring that each dimensionality class was proportionally represented in both sets. The standard split ratio of 80\% training and 20\% testing was applied. To reduce the influence of random partitioning on model performance, this process was repeated across five independent random seeds, and the reported metrics were averaged across all runs. 

Together, the cross-validation and ROC analysis confirm that our model achieves high predictive accuracy, is generalizable, and effectively mitigates the class imbalance effects that hinder minority class performance in the small HMH dataset.

\section{Discussion}
\label{discussion}

Accurately predicting the structural dimensionality of hybrid metal halides (HMHs) is highly important for the rational design of functional materials~\cite{mao_2025_a,jo_2024_comprehensive}. Especially for applications where low-dimensionality is associated with enhanced optoelectronic properties~\cite{lu_2024_machine}. One of the challenges is that the existing HMH datasets are often small and highly imbalanced. This makes the development of machine learning models much more complex. Our study shows that feature engineering and class imbalance mitigation can substantially improve the predictive accuracy while preserving interpretability.

 Models trained on original descriptors show limitations in generalization and especially in capturing the features of 0D and 1D HMHs, resulting in low F1-scores. To circumvent these challenges we have introduced interaction-based descriptors, which encode steric hindrance, bonding geometry, and coordination effects to better reflect the structure-property relationships that influence HMHs' structural dimensionality. Feature importance analysis confirms that these engineered descriptors were among the top contributors to model prediction accuracy. Especially the descriptors related to organic cation structure and hydrogen bonding, supporting observations from prior studies~\cite{lyu_2021_predictive,yuan_2023_accurate}. 

Beyond the feature design, our results highlight that addressing data imbalance is essential for model generalization and reliable prediction. Consistent with earlier findings that imbalance reduces minority class recall~\cite{fernandez_2018_smote}. We have improved the minority class performance by applying SMOTE oversampling, which led to a substantial increase in the F1-score for 0D structures from 0.29 to 0.74. The model successfully integrated the synthetic data without overfitting. ROC analysis further confirmed this trend and shows significant improvements in AUC for minority classes after augmentation.

Ensemble-based models such as random forests and gradient boosted trees proved to be highly effective in this small-data regime, aligning with relevant studies~\cite{suguna_2025_mitigating,mienye_2022_a,sagi_2018_ensemble,grinsztajn_2022_why}. The cross-validation scores are fairly high and stable (mean F1 = 0.964), which suggests that the final model generalizes well and is not overly sensitive to data splits. This is especially important given the variability in small datasets. Our framework, including the integration of interaction descriptors, class imbalance mitigation, and ensemble learning, provides the possibility of applying ML models to other underexplored HMH properties.

While SMOTE oversampling helps to mitigate the effects of class imbalance, it relies on interpolation and may not capture the full diversity of HMH structures. Future work will explore generative modeling approaches to incorporate structures for minority classes. In addition, expanding the dataset with experimentally or computationally validated structures would further improve the model's robustness. Coupling this framework with models that learn directly from compositional strings or text-based representations could reduce the reliance on handcrafted descriptors and accelerate discovery. 

Overall, these results demonstrate that domain-informed feature design, combined with data-level augmentation and ensemble learning, provides a transferable workflow for small, imbalanced materials datasets. Although demonstrated here for HMH dimensionality prediction, this approach is broadly applicable to other structure-property prediction tasks in functional materials.

\section{Conclusion {and future work}}
\label{conclusion}

In this study, we present a machine learning framework for the structural dimensionality prediction of hybrid metal halides (HMHs). We address two main challenges, namely small dataset size and significant class imbalance. By introducing the interaction-based descriptors and applying SMOTE-based oversampling, we significantly improved the prediction accuracy for minority classes (0D and 1D) without sacrificing the performance. Feature importance analysis highlighted the relevance of interaction descriptors, while ROC and cross-validation results confirm the model's robustness and generalizability. 

Our results show that combining domain-specific feature engineering with class imbalance mitigation leads to more reliable ML models for underrepresented regions of materials space. The methods presented here can be applied to broader HMH discovery efforts that go beyond the well-studied 2D structures. In addition to predictive gains, the model provided interpretable insights into the structural chemistry of HMHs. Feature importance analysis revealed that interaction features emphasizing the role of hydrogen bonding and steric interactions were strongly weighted. Collectively, these findings show that dimensionality is strongly governed by a balance between steric bulk, polarity, and hydrogen bonding, which were encoded in the interaction features. 

Future work will involve several directions to extend this study. First, we will expand our dataset by generating more computational and experimental data to get a more in-depth understanding of the property-structure relationship. Additionally, large language models (LLMs) represent a rapidly emerging tool for materials research. Future work will involve training and fine-tuning LLMs to assist in structural dimensionality analysis by leveraging textual descriptors of chemical formulas. Together, these strategies can advance the development of data-driven tools for novel and underexplored materials.

\section{Supplementary Information}

\begin{figure}
    \centering
    \includegraphics[width=1.0\linewidth]{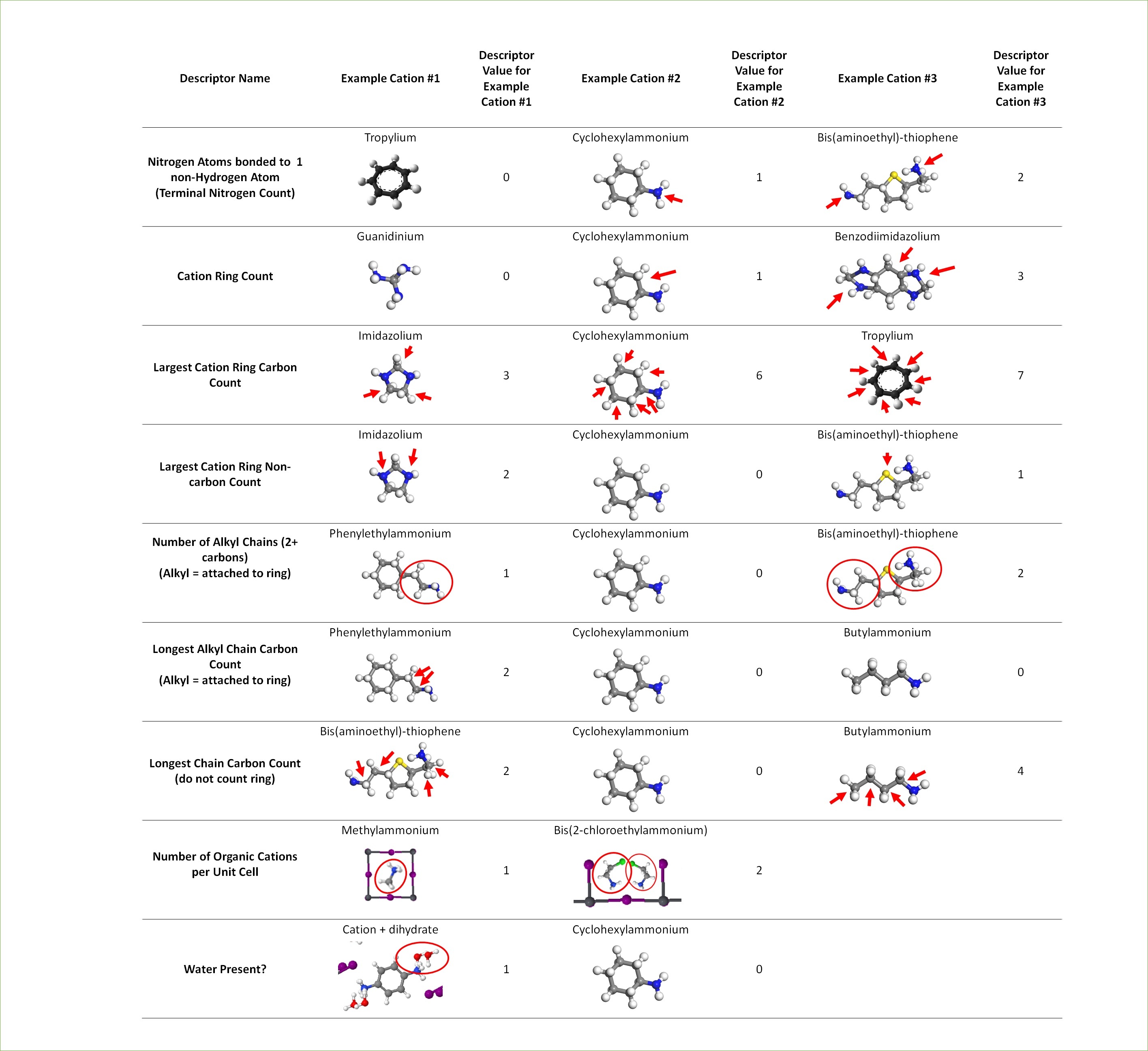}
    \caption{Original features. These chemical descriptors and the formulas for the organic and inorganic components provided the initial features for training our ML model. These features sought to capture key cation structural traits in a simple description.  }
    \label{fig:original_features}
\end{figure}

This section provides an in-depth description of the original features used in this work (Figure~\ref{fig:original_features}). These features were derived from the HybriD\textsuperscript{3} dataset and capture structural and chemical properties of HMHs. Each descriptor encodes a specific aspect of the organic cation or inorganic framework that influences the resulting structural dimensionality. The first two descriptors (not shown in Figure~\ref{fig:original_features}) are text strings of the organic and inorganic formula units, respectively. For example, for HMH formula C\textsubscript{8}H\textsubscript{8}N\textsubscript{4}PbI\textsubscript{4}, the organic text string descriptor would be C8H8N4, and the inorganic would be PbI4. These formula descriptors contain compositional information such as atomic identities and similarities in available lattice volume. While HMH total formulas do contain the number of duplicate cations within them (i.e. "2" in (C\textsubscript{6}H\textsubscript{14}N)\textsubscript{2}PbI\textsubscript{4}), this information was not always included in organic formulas in HybriD\textsuperscript{3}, so the number of duplicate cations was included as another descriptor. The terminal nitrogen count descriptor determines the number of nitrogen atoms bonded to a single non-hydrogen atom. 
N was chosen for this descriptor due to its prevalence in the dataset and importance to overall charge distributions within the organic cations. For instance, a cation with three terminal N will apply stronger forces on the surrounding inorganic lattice when compared to a cation of similar size with only a single terminal N, possibly distorting the inorganic structure and changing dimensionality of the HMH. The number of cation rings descriptor corresponds to the number of rings within the organic cation, which can affect steric packing and $\pi - \pi$ interactions that may influence dimensionality. The number of carbon vs. non-carbon atoms within the largest rings were also tracked as separate descriptors to provide more information on local charge. Number of alkyl chains captures the aspects of molecular shape and steric hindrance introduced by side chains. Relative length of the alkyl chains, and carbon number within the longest chain, also sought to simply capture alkyl chain contributions. Finally, water presence may affect stability and is a Boolean type descriptor. In total, these simple descriptors captured major geometric details of cation structures and proved to be a successful basis for the creation of interaction features for our ML model.


\section*{Acknowledgements}

This work has been supported by the National Science Foundation grants DMREF 2323546, DMREF 2323547, and DMREF 2323548, and used resources of the Oak Ridge Leadership Computing Facility at the Oak Ridge National Laboratory, which is supported by the Office of Science of the U.S. Department of Energy under Contract No. DE-AC05-00OR22725.

\section*{Data Access Statement}

The HMHs dataset used in this work is available at the HybriD\textsuperscript{3} database: https://materials.hybrid3.duke.edu

\section*{Conflict of Interest Statement}
The authors do not have conflicts of interest to declare. 

\section*{Ethics Statement}
No confidential information about human subjects is used in this manuscript.

\bibliographystyle{unsrt}
\bibliography{main}

\end{document}